\documentclass{article}
\usepackage{ijcai16}

\usepackage{times}

\usepackage[latin9]{inputenc}
\usepackage{color}
\usepackage{array}
\usepackage{booktabs}
\usepackage{multirow}
\usepackage{amsmath}
\usepackage{amssymb}
\usepackage{graphicx}
\usepackage{amsmath,amssymb} 
\usepackage{color}
\usepackage{array}
\usepackage{multirow}
\usepackage{graphicx}
\usepackage{color}
\usepackage{times}
\usepackage{epsfig}
\usepackage{tabularx}
\usepackage{array}
\usepackage{algorithmic}

\usepackage[ruled,noend]{algorithm2e}
\usepackage[T1]{fontenc}
\usepackage{color}
\usepackage{booktabs}
\usepackage{amstext}
\usepackage{relsize}
\usepackage{lipsum}
\usepackage{array}
\usepackage{slashbox}

\mathchardef\mhyphen="2D 

\newcommand{\thickhline}{%
    \noalign {\ifnum 0=`}\fi \hrule height 1pt
    \futurelet \reserved@a \@xhline
}
\newcolumntype{"}{@{\hskip\tabcolsep\vrule width 1pt\hskip\tabcolsep}}
\makeatother

\newlength{\Oldarrayrulewidth}

\newcommand{\qed}{\nobreak \ifvmode \relax \else
      \ifdim\lastskip<1.5em \hskip-\lastskip
      \hskip1.5em plus0em minus0.5em \fi \nobreak
      \vrule height0.75em width0.5em depth0.25em\fi}

\begin{document}
\title{CS591 Report: Application of siamesa network in 2D transformation}
\author{Dorothy Chang\\
Department of Computer Science\\
University of Navarra\\
dchang@unav.es}
\maketitle

\begin{abstract}
Deep learning has been extensively used various aspects of computer vision area. Deep learning separate itself from traditional neural network by having a much deeper and complicated network layers in its network structures. Traditionally, deep neural network is abundantly used in computer vision tasks including classification and detection and has achieve remarkable success and set up a new state of the art results in these fields. Instead of using neural network for vision recognition and detection. I will show the ability of neural network to do image registration, synthesis of images and image retrieval in this report. 
\end{abstract}

\section{Part One: Image Retrieval}
\label{sec:Part1}
I work with Di for part of my time this semester to help him develop a neural network that can tell whether two images are identical or not. We further extend this network to one which after being well trained will help image retrieval task.

\subsection{Introduction}
Most existing image matching methods are based on features, like color histogram\cite{zhang2014learning}\cite{zhang2015learning1}\cite{zhang2012learning}, SIFT, HOG, etc. 
However, those features which are good for natural image matching fail on the same matching task 
of document images. It is determined by the characterises of document images\cite{zang2015learning}. Most document images are 
binarized black white images. Features extracted based on color are not suitable for processing 
document images. Features extracted based on gradient are not desirable as well, since intensity gradient is not strong on document images. There are some methods designed for document image matching, including feature based\cite{zhang20083d}\cite{zhang2015learning}, structure based and OCR based\cite{zhang2016cgmos}. But they all have limitations. This part will be discussed in the related works section. The proposed method uses CNN to achieve document image matching. It has a better performance on matching document images in various conditions than existing methods.
Furthermore, a specific metric usually is required\cite{zhang2014alignment} when computing a distance between pairs of extracted features. Such metric is often chosen heuristically or empirically. Using these metrics in retrieval tasks though may achieve good results, the solution is only suboptimal. In the proposed method, instead of choosing metric ourself, a neural network is designed and employed to automatically measure distance between two feature vectors. This is one of primary contribution of this work.

There three novel contribution in this work: (1) a neural network composing two neural networks is proposed to solve document image matching problems. The proposed approach is much better insistent to occlusion, spacial transformation and other types of noises. (2) A distance computation mechanism is integrated in the proposed system that distance similarities between pairs of images can be automatically handled by the system, which is demonstarted to be more robust and accurate than existing methods.

\begin{figure*}[ht]
\centerline{ %
\begin{tabular}{c}
\resizebox{1\textwidth}{!}{\rotatebox{0}{ 
\includegraphics{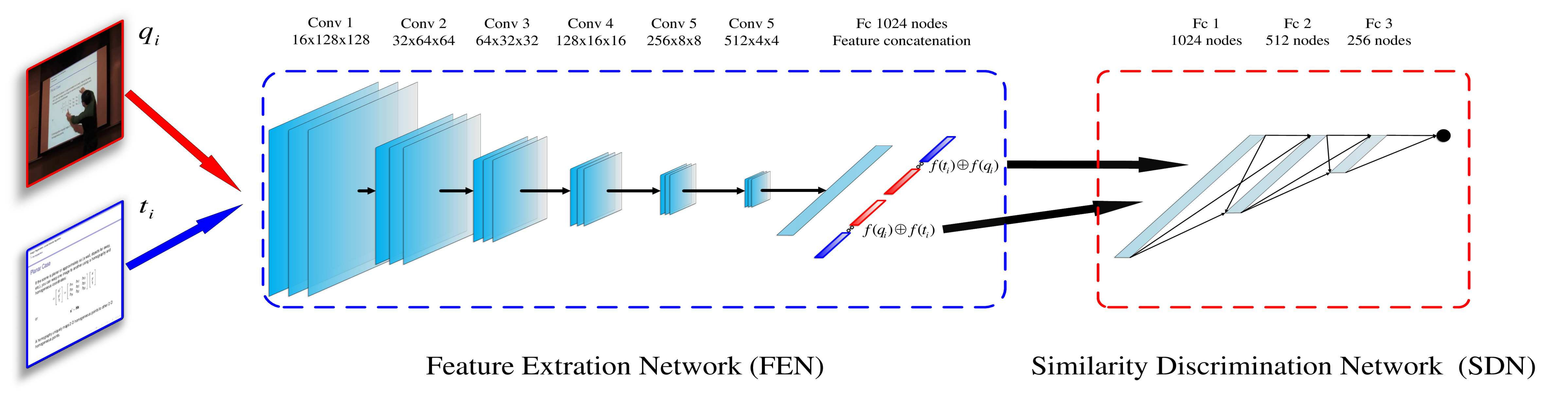}}}
\end{tabular}} 
\caption{The architecture of the proposed neural network system. For convolutional layers, the receptive
fields are denoted using the conv$<depth>\times<width>\times<height>$ format.}
\label{fig_flow}
\end{figure*}

\subsection{The Proposed Approach}
In this work, a system based on deep neural network for image retrieval is designed and implemented. In most existing deep neural networks, network is used only for feature extraction and a pre-defined metric then is heuristically or empirically chosen to measure distances between image pairs based on extracted features. Although good image retrieval results can be achieved by using existing framework, the results usually are sub-optimal and it is sometime tricky when choosing appropriate distance metric.  

To enable automatically feature extraction and similarity computation, the proposed system composes two neural networks that are connected with each other: a feature extraction network (FEN) and a similarity discrimination network (SDN). The FEN extracts features from input images and then the SDN automatically computes a similarity score between a pair of images features. The two networks are trained as a whole system in an end-to-end manner that to train the system, given training data as query images $Q = \{q_i\}$ and target images $T = \{t_i\}$, and there is at least one matching image for each $q_i\in Q$ in $T$. We then concern the training of the system as a supervised training problem that for each training sample $(x_i, y_i)$, $x_i$ contains a pair of images $(q_i, t_i)$ randomly chosen from $Q$ and $T$, and $y_i$ is a $\{0, 1\}$ binary label representing whether the pair of images $(q_i, t_i)$ is a match. During training, we fed image $q_i$ and $t_i$ to FEN to extract their features $f(q_i)$ and $f(t_i)$ respectively. The features $f(q_i)$ and $f(t_i)$ are then concatenated to form a longer feature vector $f(q_i)\oplus f(t_i)$. 

The order of $f(q_i)$ and $f(t_i)$ in the concatenation determines how SDN is trained. Given any pair of image features, SDN should be order-invariant. Thus, We also create a symmetric feature vector  $f(t_i)\oplus f(q_i)$ and fed it to SDN as well. An average error of these the two feature vectors is computed. We actually notice that he retrieval results can be enhanced if symmetric concatenation can be employed. We believe this is because that when SDN is trained in a order-invariant manner, features extracted using FEN are more robust and representative when used to distinguish a pair of images.

At the output end of the system, we penalize a cross entropy error between prediction $\hat{y}_i$ and $y_i$ using following equation:

\begin{equation}
E_i = - y_i \log(\hat{y}_i) - (1-y_i)\cdot\log(1-\hat{y}_i)
\end{equation}

In this architecture of neural network, FEN is not trained individually and we do not penalize errors of features generated by FEN directly. This is because the goal of the entire system is to compute a similarity measure for any given pair of images. For each single image fed to FEN, it is hard to define an appropriate objective function to compute a lost at the output end of FEN. Thus FEN is co-trained with SDN, and FEN is optimized when gradient flow back propagate from FEN. 

Indeed, optimizing one network using another network is not uncommon. Similar to the mechanism of the proposed neural network, in adversarial network \cite{goodfellow2014generative} to improve qualities of images generated by a generator network, only the cost of a discriminator network is used to optimize entire system. Neural network optimized implicitly using another network usually achieve a better performance, although people are still unclear about the exact mechanism and reasons behind this phenomenon, it turns out that the neural network actually is "smart" enough to know how to generate an output that performs the best.

The proposed system contains two networks. A feature extraction network (FEN) and a similarity discrimination network (SDN). The architecture of these two networks are shown in \ref{fig_flow}.

The FEN is a 8 layers network consisting of 7 convolution layers and 1 fully connected layer. Batch normalization \cite{ioffe2015batch} followed by a leaky rectified linear unit (LReLU) are used for all convolution layers in FEN. Such configuration has been demonstrate both efficient and effective in many literatures \cite{kare,kriz,hint}. Kernels of $3\times 3$ size are used universally. There are 16 kernels at the first convolution layer. We double the number of kernels at each new layer, in the meantime, the sizes of images are halved using max-pooling. A full connection layer with 1024 nodes is stack on top of the last convolution layer. A tanh activation is used for this layer. 

The SDN is a 3-layer fully connected neural network. The number of neurons in each layer are: 1024, 512 and 256. tanh is used for all three layers. We scale the output value of the SDN to $[0, 1]$.

\subsection{Experiment Results}
\begin{figure*}[ht]
\centerline{ %
\begin{tabular}{c}
\resizebox{0.7\textwidth}{!}{\rotatebox{0}{ 
\includegraphics{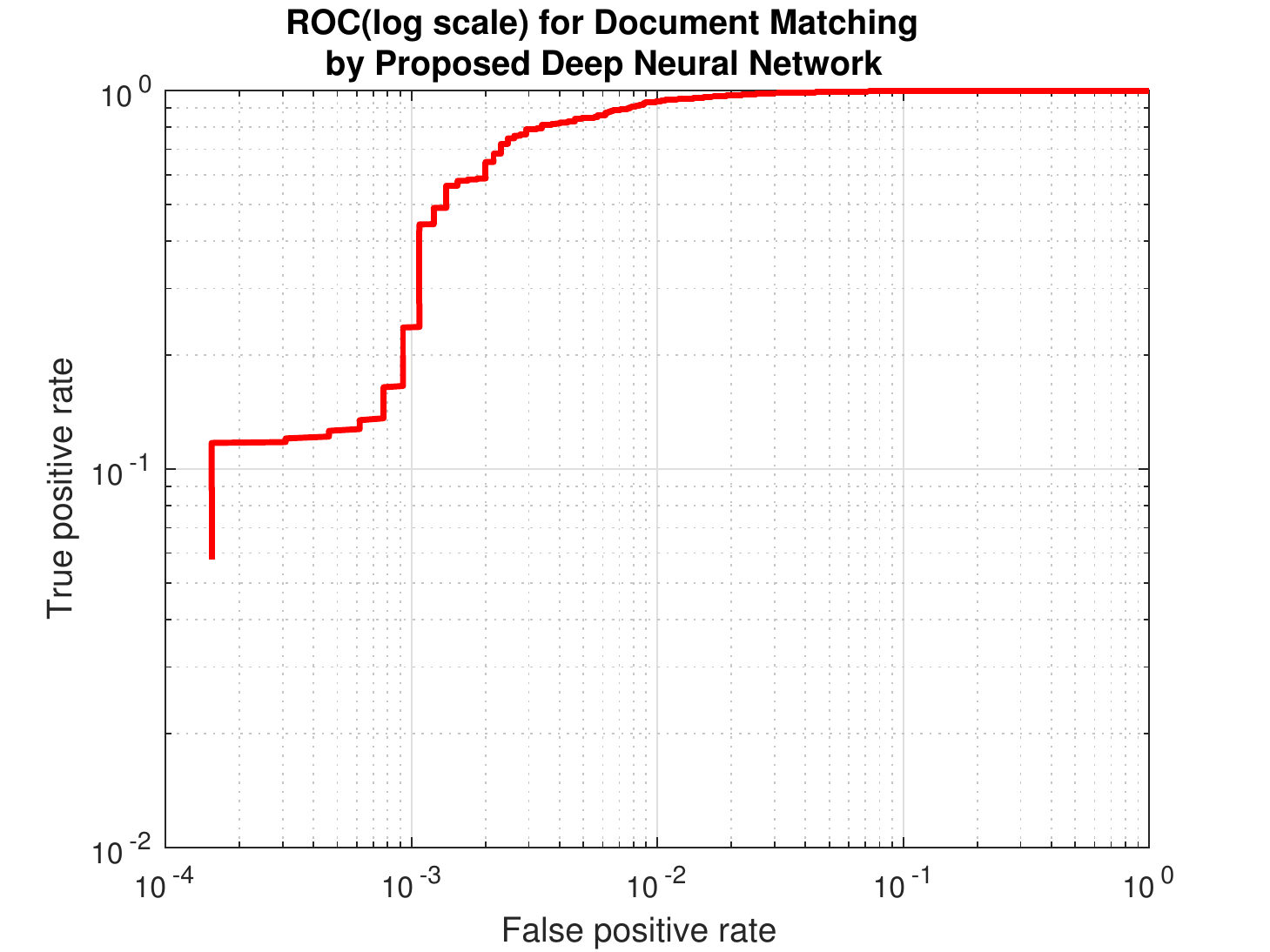}}}
\end{tabular}} 
\caption{The performance of maching network.}
\label{fig_cnn_performance}
\end{figure*}

We evaluate the proposed deep neural network by using 13000 document image pairs. 6500 of them are correct image pairs with same content. The other 6500 of them are image pairs with different content. Figure \ref{fig_cnn_performance} shows the performance of document matching by the proposed deep neural network. The ROC curve of the proposed method indicates the proposed method has a very good performance for distinguishing if two document images have same content.

We also evaluate the proposed deep neural network method for document image retrieval.
The 6500 original document images are consist of the target images. The 6500 warped noisy document images are query images. The proposed deep neural network could generate a likelihood value for each image pair. Therefore, we can sort the likelihood and pick out the best candidates of matched document images from the target image set. We compare the proposed method with the benchmark method \cite{vita}. Top 1, top 5 and top 10 hit rates are shown in Table  \ref{table_compare}.

\begin{table}
\begin{tabular}{|c||c|c|c||c|c|c|}
\hline 
noise & b@1 & b@5 & b@10 & p@1 & p@5 & p@10 \\ 
\hline 
0 & 0.846 & 0.961 & 0.986 & 0.882 & 0.981 & 0.981 \\ 
\hline 
1 & 0.813 & 0.936 & 0.967 & 0.913 & 1.0 & 1.0 \\ 
\hline 
2 & 0.747 & 0.892 & 0.934 & 0.888 & 0.969 & 0.993 \\ 
\hline 
3 & 0.664 & 0.826 & 0.884 & 0.888 & 0.987 & 0.993 \\ 
\hline 
4 & 0.543 & 0.752 & 0.826 & 0.851 & 0.981 & 0.993 \\ 
\hline 
5 & 0.458 & 0.681 & 0.782 & 0.771 & 0.932 & 0.938 \\ 
\hline 
6 & 0.436 & 0.642 & 0.744 & 0.765 & 0.913 & 0.950 \\ 
\hline 
7 & 0.456 & 0.634 & 0.730 & 0.740 & 0.913 & 0.944 \\ 
\hline 
8 & 0.414 & 0.596 & 0.678 & 0.648 & 0.870 & 0.913 \\ 
\hline 
9 & 0.376 & 0.596 & 0.689 & 0.592 & 0.790 & 0.858 \\ 
\hline 
\end{tabular} 
\caption{The hit rates of the proposed method and the benchmark on image retrieval with different noise level.(b is for benchmark, p is for proposed method. @1, @5 and @10 are for top 1, 5 and 10 cases.)}
\label{table_compare}
\end{table}

\section{Part Two: Image Registration}
Image registration is a fundamental problem in many applications. Generally, methods of registration can be grouped to two families. The first one is intensity based registration method, where images are registered by iteratively minimizing intensity difference between source image and target image. The second one is feature point based registration, where feature points are detected in both source and target images. Then the registration is solved by computing the transformation between the statistically best corresponding point pairs in two images. 

\begin{figure*}[ht]
\centerline{ %
\begin{tabular}{c}
\resizebox{1\textwidth}{!}{\rotatebox{0}{ 
\includegraphics{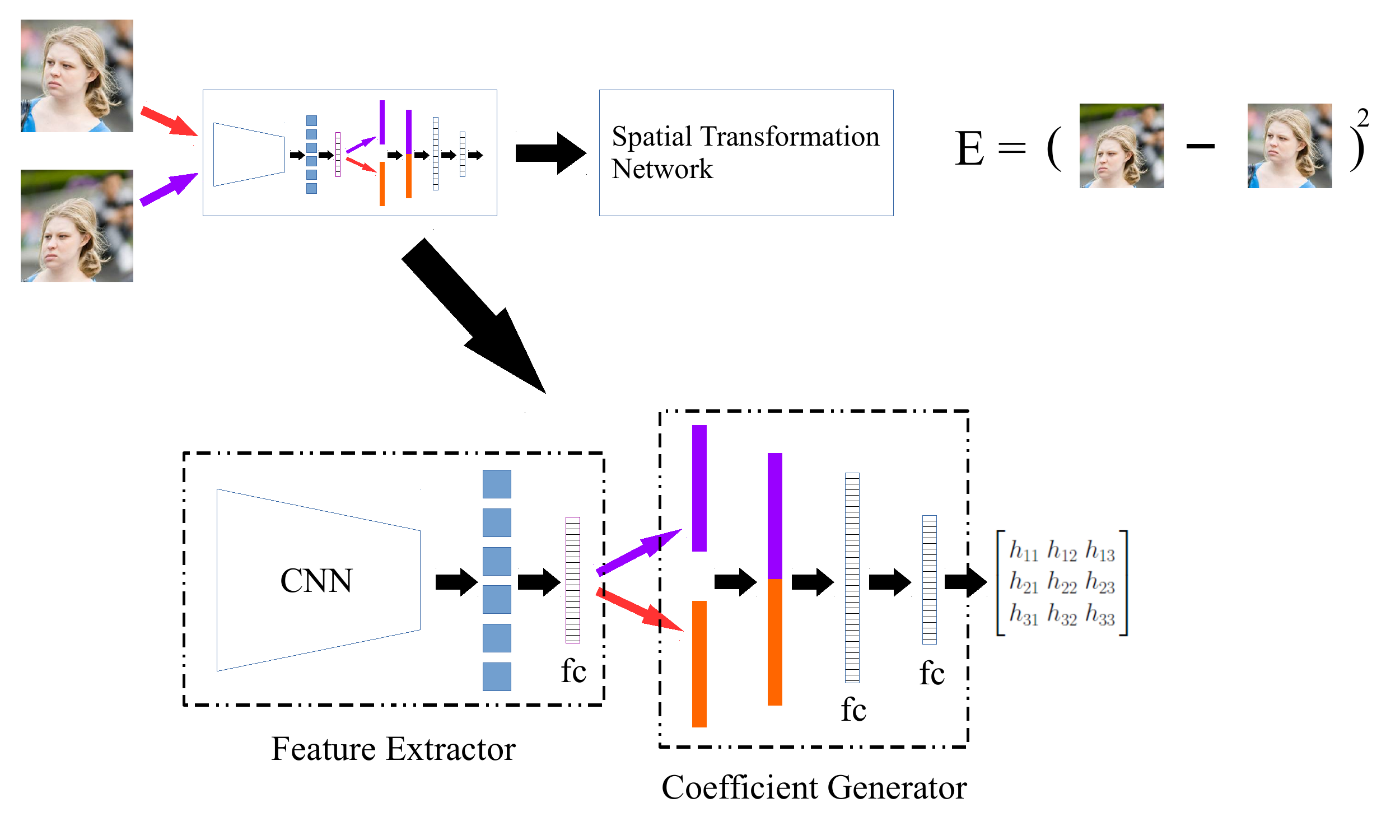}}}
\end{tabular}} 
\caption{The structure of the proposed registration network.}
\label{fig_RegNetStruct}
\end{figure*}

Almost all existing registration algorithms treat image registration individually for each case. There is no knowledge sharing among different images. Different from traditional registration algorithm, I proposed a framework using deep neural network which can register images in one shot.

\subsection{The Proposed Registration Network} 
Registration using neural network has not drawn many attention in the past. There are just a few works which utilize neural network into their registration algorithm \cite{DBLP:journals/corr/MiaoWL15}\cite{sabisch1998automatic}. A common feature of neural networks in these existing works is that the networks are always trained supervisely in the sense that the network is trained as a regressor which fits output to transformation parameters.

Instead of training the network supervisely using known transformation parameters in training data, in the proposed neural network, we do not use any ground truth transformation parameters during training. Rather than calling the proposed network a regressor, I would prefer to call it a inferencer in the sense that all transformation parameters are inferred by the network automatically during training.

The ability the proposed network can automatically infer transformation parameters comes from a spatial transformation module (STM) \cite{jaderberg2015spatial} we used in the proposed network. Different from the rest part of the network, STM is only responsible for applying spatial transformation in forward passing. There is no gradient passing through during back propagation. The detailed structure of the proposed neural network is shown in Figure \ref{fig_RegNetStruct}. Besides the STM aforementioned, the primary part of the proposed network is a parameter inference module (PIM). Inside the PIM, there are two connected networks responsible for different jobs. The first network is a feature extractor which is basically a convolutional neural network and can be composed by any main stream CNN structure. The second network is a coefficient inferencer which generates transformation parameters given a concatenated feature extracted by feature extractor for a image pair.

In current implementation, only linear transformation is implemented. However, by changing the implementation of STM, the proposed neural network can be easily extended to tackle nonlinear transformation as well.

I have tried many different CNN structures in feature extractor network including VGG\cite{simonyan2014very}, NIN\cite{lin2013network} and DCNN\cite{huang2016densely}. Deeper network will make convergence faster and a slightly better registration result. 

\subsection{Experiment Results}
To evaluate the proposed network on image registration task, I randomly download 48,000 images from flickr and artificially added transformation to downloaded images. Inside these 48,000 images, 45,000 images will be used to train the network and we use remaining 3,000 image to test.

We use two existing registration algorithm as our baseline. The first one is feature point base registration algorithm using SIFT as features. The second one is a intensity based registration algorithm called ECC \cite{evangelidis2008parametric}. Generally speaking.

There are still a lot of experiments going on these days. I will only provide some preliminary qualitative results in this report. Generally speaking, for easy dataset, SIFT performs the best, and ECC is slightly better than the proposed approach. However, when image is blurred and polluted by noises, SIFT based algorithm and ECC will fail very often. In this case quantitatively, the proposed approach performs better.

Qualitative results of the proposed approach tested on easy dataset and hard dataset area shown in Figure \ref{fig_easy} and Figure \ref{fig_hard} respectively.

\begin{figure*}[ht]
\centerline{ %
\begin{tabular}{c}
\resizebox{1\textwidth}{!}{\rotatebox{0}{ 
\includegraphics{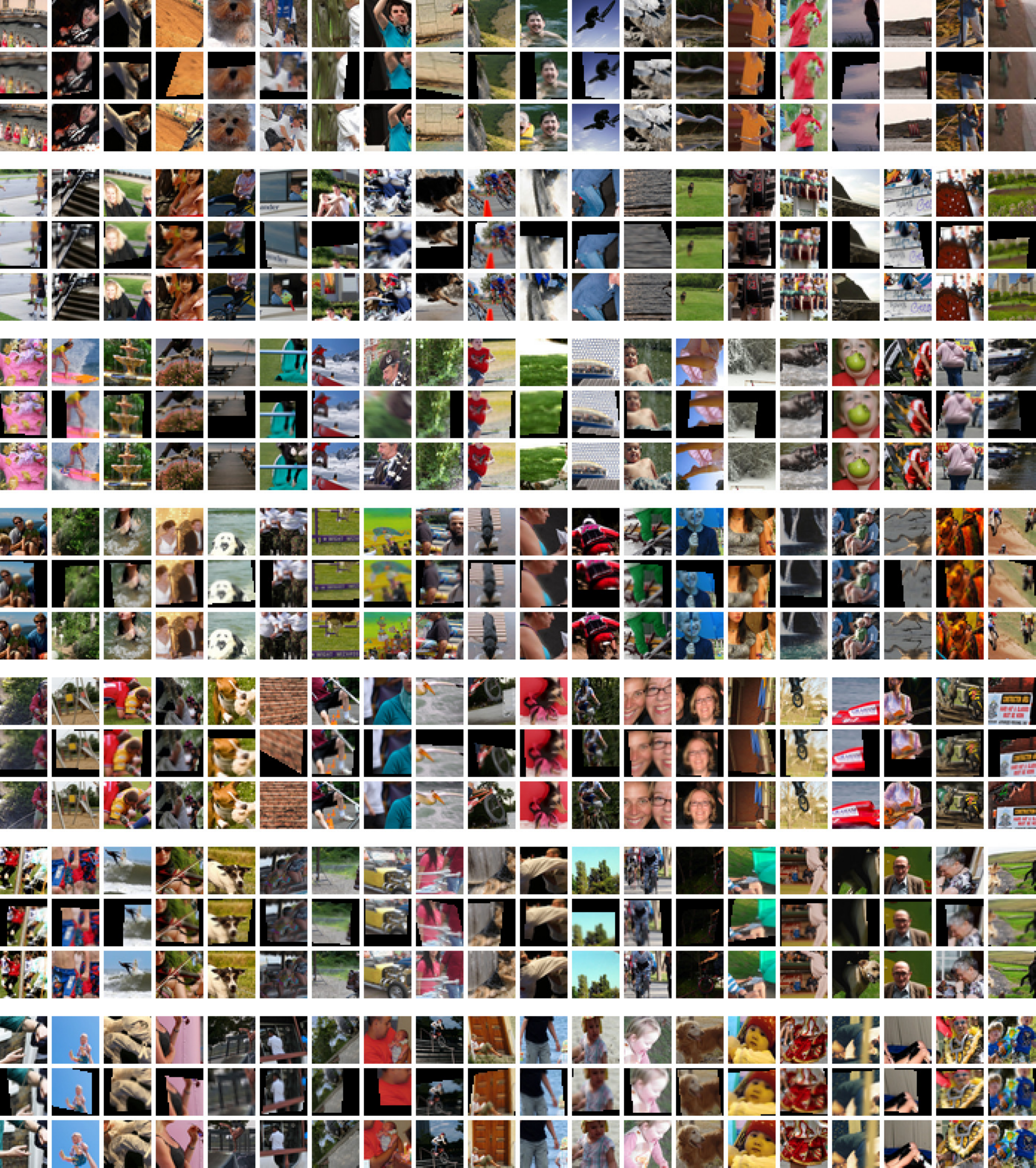}}}
\end{tabular}} 
\caption{Results of registration using the proposed approach on easy dataset.}
\label{fig_easy}\end{figure*}

\begin{figure*}[ht]
\centerline{ %
\begin{tabular}{c}
\resizebox{1\textwidth}{!}{\rotatebox{0}{ 
\includegraphics{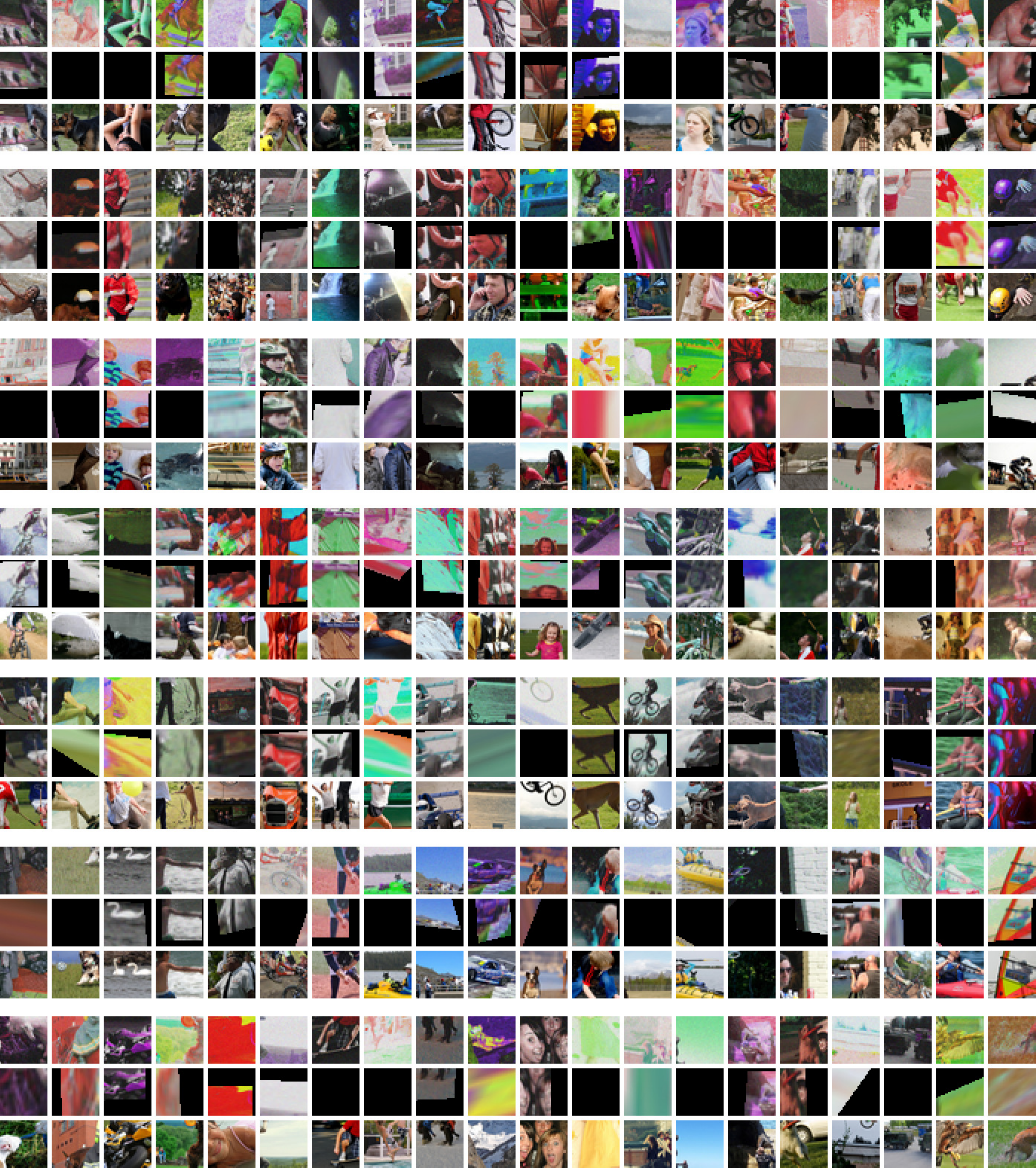}}}
\end{tabular}} 
\caption{Results of registration using the proposed approach on hard dataset.}
\label{fig_easy}\end{figure*}

\bibliographystyle{named}
\bibliography{mybib}

\end{document}